\documentclass{article}

\usepackage{arxiv}

\usepackage[T1]{fontenc}    
\usepackage{hyperref}       
\usepackage{url}            
\usepackage{booktabs}       
\usepackage{amsfonts}       
\usepackage{nicefrac}       
\usepackage{microtype}      
\usepackage{lipsum}
\usepackage{graphicx}
\title{Automatic Generation of Chinese Handwriting via Fonts Style Representation Learning}

\author{
  Fenxi Xiao \\
  Department of Electrical Engineering\\
  Jinan University\\
  Guangzhou, China 510632 \\
  \texttt{xfx0203@163.com} \\
  \And
  Bo Huang \\
  Department of Electrical Engineering\\
  Jinan University\\
  Guangzhou, China 510632 \\
  \texttt{abohuang@jnu.edu.cn} \\
   \And
 Xia Wu \\
  Department of Electrical Engineering\\
  Jinan University\\
  Guangzhou, China 510632 \\
  \texttt{wuxia\_liao@qq.com} \\  
}

\begin{document}
\maketitle

\begin{abstract}
  In this paper, we propose and end-to-end deep Chinese font generation system.
  This system can generate new style fonts by interpolation of 
  latent style-related embeding variables that could achieve 
  smooth transition between different style. 
  Our method is simpler and more effective than other methods,
  which will help to improve the font design efficiency.
\end{abstract}

\keywords{Chinese Fonts \and Generation \and Handwriting \and Font Style \and Representation Learning}

\section{Introduction}

Language is a unique symbol of human civilization. Writing system is the development of the language.
Writing system usually has many different fonts.  Unlike the English font library 
that has only 26 alphabets, 
Chinese font library contains tens of thousands of characters. There are also thousands of commonly 
used characters. Making a new style font library is a difficult and time-consuming job due to the 
huge amount of Chinese characters. Designing a new font of Chinese character 
require a lot of human design and adjustment to draw each character.
It is necessary to find a automation method to help the font designing 
for Chinese character.

In the last few years, the development of deep learning make it possiable in automatic 
image style transfer. Several researcher have intended to generate Chinese fonts by using
different deep learning method. Jiang et.al. using a U-net model realize end-to-end
Chinese character mapping, that automatically generate the whole GB2312 font library
 that consists of 6763 Chinese characters from a small number of characters written by the user.
\cite{Jiang2017}. Jiang develop an efficient and generalized deep framework W-Net, 
that is capable of learning and generating any arbitrary characters 
sharing the style similar to the given single font character\cite{amari_w-net_2017}.
Sun et. al. also propose a variational auto-encoder framework 
to generate Chinese characters\cite{Sun2017}.

These methods are based on such an assumption that the latent features
of a Chinese character can be disentangling 
into content-related and style-related components. 
Combining different parts content-related and style-related components 
can construct new style fonts. 
In this paper, we propose and end-to-end deep Chinese font generation system.
This system can generate new style fonts by interpolation of 
latent style-related embeding variables that could achieve 
smooth transition between different style. 
Our method is simpler and more effective than other methods,
which will help to improve the font design efficiency\cite{Sun2017,Lyu2018,Deng,guo_creating_2018,
chang_rewrite2_nodate,chang_generating_nodate,azadi_multi-content_nodate}.

\section{Method Description}
Due to the complicated structures of Chinese characters, It is not easy transfer 
deep learning methods widely used in image synthesis to this job.
The big problem is that the style and content features of Chinese characters are 
complexly entangled. The present deep learning methods, such as cross-domain 
disentanglement\cite{Gonzalez-Garcia2018}, Dientangled Representation\cite{Lee2018},
U-Net model\cite{esser_variational_nodate}, are not well work on Chinese characters synthesis.

In this paper, we using a encoder-decoder model to map Chines character with one type to 
another type, such as Song fonts to Kai fonts. The mapping just transfer the character style,
the characters have the same content. During the mapping training, the encoder-decoder model 
will extract the fonts style features in the latent space. When we find the right model parameters,
we embeding a font style one-hot vector in the latent space. Then we retrain the model with 
style embding. This process can be seen as an artificial entanglement process,that entangled 
the style and content features of the characters.

Our method contains three steps. First, we using a U-Net encoder-decoder model the extract the 
font feature vectors for about 40 different type fonts. Then, we concatenate an one-hot vector 
to the feature vectors, and retrain the model. The one-hot vector embding  about 40 different fonts
style. Finally, we can change the one-hot style vector to arbitrary embedding vectors, and synthesis 
new style fonts which have mixed styles of different fonts。

\subsection{U-net encoder-decoder model}

\begin{figure}
  \centering
  \includegraphics[scale=0.5]{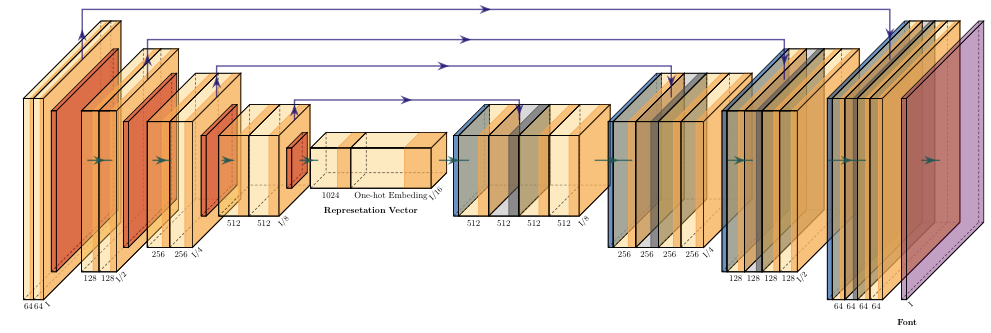}
  \caption{U-Net encoder-decoder architecture}
  \label{fig1}
\end{figure}

The U-net encoder-decoder architecture is shown in Figure \ref{fig1}. The input
of the network is an Song style Font with 256x256 pixels. The Font throught and
8-layes convolutional encoder. The output of the encoder is 5x5x512 features 
vectors. This features  concatenate a one-hot style 
embedding are input to the decoder. After similar 8-layers deconvolution, the 
decoder output a new style fonts.

\section{Experiments results}
\begin{figure}
  \centering
  \includegraphics[scale=0.9]{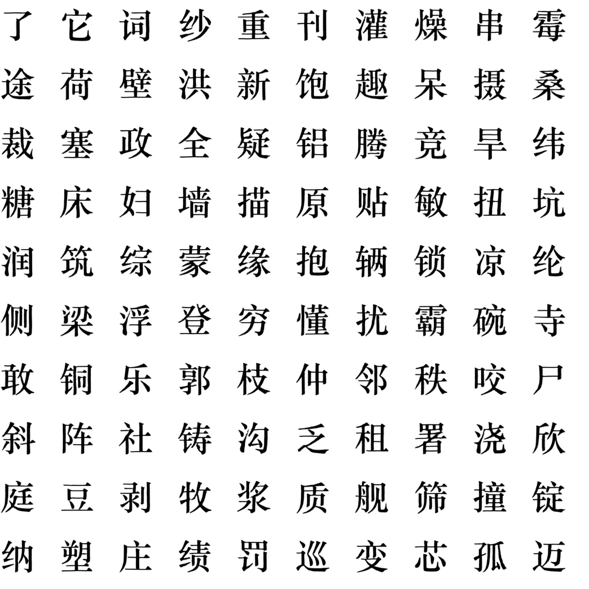}
  \caption{Hei style fonts generation}
  \label{fig2}
\end{figure}

\begin{figure}
  \centering
  \includegraphics[scale=0.9]{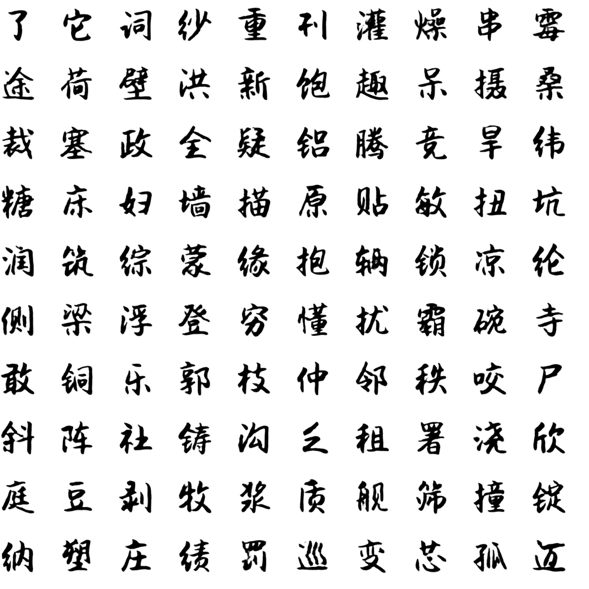}
  \caption{Xing style fonts generation}
  \label{fig3}
\end{figure}

Figure \ref{fig2}\ref{fig3} is the two different output of the model.
The input is all the Hei style  fonts. Figure \ref{fig2} is the  Song 
style fonts. Song style have a similar style feature of Hei. Figure \ref{fig3}
is XingKai style, that has very different characteristics style with the Hei style.
This two output examples  show that the model has a good ability to generate 
new fonts. 

\begin{figure}
  \centering
  \includegraphics[scale=0.8]{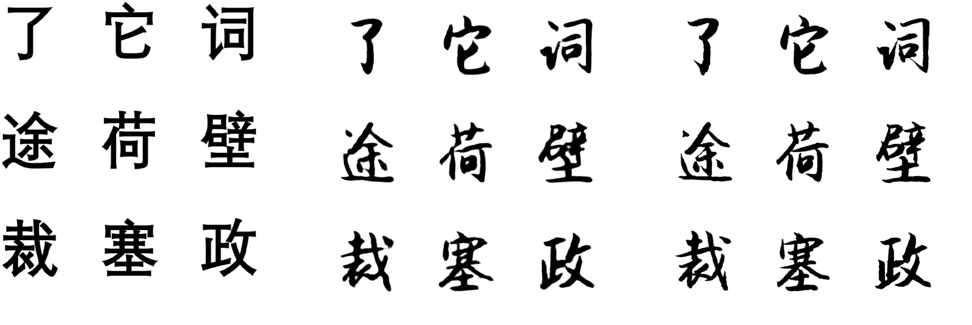}
  \caption{Hei style to Xing style transfer}
  \label{fig4}
\end{figure}

The style transfer have been show in Figure\ref{fig4}. The first three 
columns are the input fonts. The middle three colums is the output of our 
model. The last three colums is the style fonts, which features is embedding 
in the one-hot vectors.  This results  indicate that our model can control 
the output fonts through the one-hot embedding vectors.

\begin{figure}
  \centering
  \includegraphics[scale=0.8]{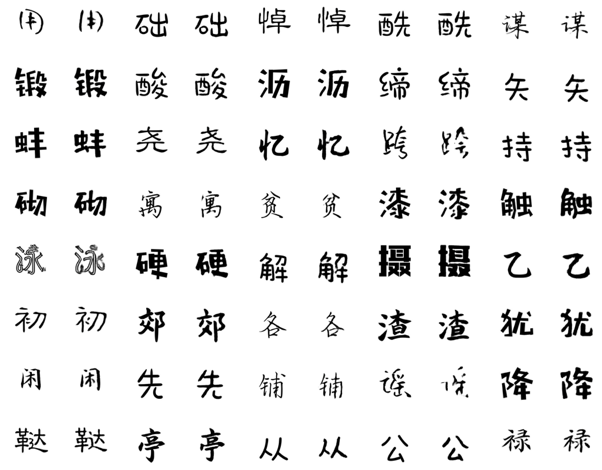}
  \caption{multiple different styles generation}
  \label{fig5}
\end{figure}

Figure \ref{fig5} show the generate fonts 
of multiple different styles. All
 the generated fonts have no missing strokes, and the font 
feature details are perfect. The input fonts is all Hei style
fonts for  those multiple generation fonts. These multiple
generation fonts show that the control of 
the style embedding vector is very good. 

Since the above results show the one-hot embedding vector already 
have stable control ability over the generation fonts, we can explore
whether the embedding vector can be used to achieve the style transition
between two fonts.

\begin{figure}
  \centering
  \includegraphics[scale=0.8]{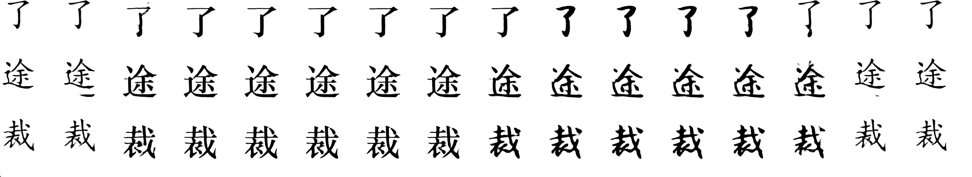}
  \caption{Control effect of one-hot embedding vector}
  \label{fig6}
\end{figure}

When the input embedding label is [1,0,...], the network will convert 
the source input font text sample to the target font text sample 
with label 0 during training, and if the the input embedding label is 
[0,1,0,..], the source font sample is converted into the 
target font sample with the label 1. So does the input embedding label
[0.5,0.5,0,...] mean that the output style is the mixture of the two fonts?
Figure \ref{fig6} show the generation of the fonts when the embedding labels are
assigned different valuse, such as [0.2,0.5,0.7,...]. The 
results indicate that the font styles can be controlled through assigned 
different valuse in the embedding vector.

\subsection{Conclusion}
Our paper proposed a novel and simple method to automatic generate new
Chinese fonts from existing font libraries. The results demonstrated that
our method is capable of generating new high-quality fonts. 

\subsection{Acknowledgement}
This work was supported by National Natural Science Foundation of China (61307080)
\bibliographystyle{unsrt}  

\bibliography{/Users/booq/Research/paper_by_me/zotero_bibtex}

\end{document}